%% file: agilab.tex
\documentclass[10pt,letterpaper]{article}
\usepackage[T1]{fontenc}
\PassOptionsToPackage{table}{xcolor}

\usepackage[pagenumbers]{agilab}
\renewenvironment{figure*}[1][t]{\begin{figure}[!htbp]}{\end{figure}}
\renewenvironment{table*}[1][t]{\begin{table}[!htbp]}{\end{table}}

\input{math_commands.tex}

\usepackage{pifont}
\usepackage{tcolorbox}
\usepackage[pagebackref,breaklinks,colorlinks,allcolors=wagiblue]{hyperref}

\newcommand{\method}{World in World}
\newcommand{\papertitle}{\method: Explore the World with World Models}
\title{\papertitle}
\author{Chenxi Song, Yanming Yang, Chi Zhang}
\wagilogo{\includegraphics[height=0.27in]{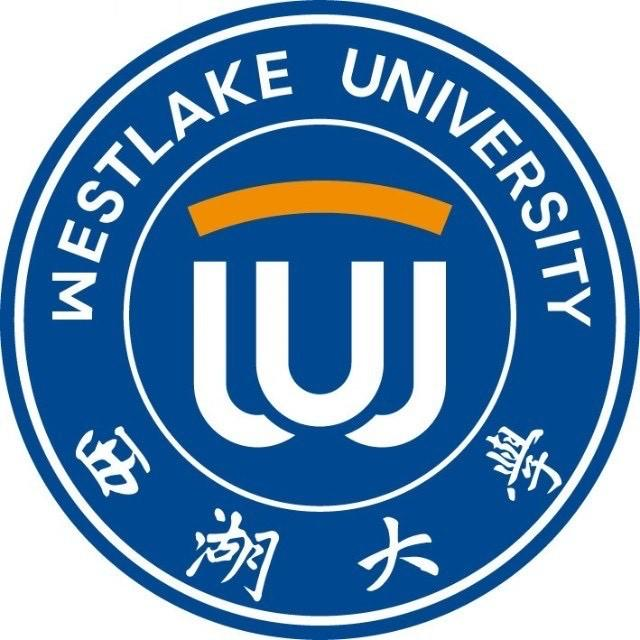}}

\hypersetup{
  pdftitle={World in World: Explore the World with World Models},
  pdfauthor={Chenxi Song, Yanming Yang, Chi Zhang}
}

\begin{document}

\vspace*{-0.50in}%
\wagibanner
\vspace{6pt}

\begin{tcolorbox}[
  colback=wagiblue!4,
  colframe=wagiblue!40,
  boxrule=0.6pt,
  arc=10pt,
  left=22pt,
  right=22pt,
  top=6pt,
  bottom=6pt,
  before skip=0pt,
  after skip=0pt
]
  {\centering
    {\LARGE\bfseries \papertitle\par}
    \vspace{8pt}
    {\large
      Chenxi Song\textsuperscript{*}\qquad
      Yanming Yang\textsuperscript{*}\qquad
      Chi Zhang\textsuperscript{\ensuremath{\dagger}}\par}
    \vspace{4pt}
    {\normalsize AGI Lab, Westlake University\par}
    \vspace{4pt}
    {\large
      \href{https://chenxi-song.github.io/worldinworld}{%
        \textcolor{red}{\texttt{https://chenxi-song.github.io/worldinworld}}}\par}
  }
  \vspace{4pt}
  \centerline{\large\bfseries Abstract}
  \vspace{4pt}
  {\itshape\noindent\input{sec/abstract}\par}
\end{tcolorbox}

\begingroup
  \renewcommand{\thefootnote}{}
  \footnotetext[0]{%
    \textsuperscript{*}\,Equal contribution.\quad
    \textsuperscript{\ensuremath{\dagger}}\,Corresponding author.}
\endgroup
\vspace{4pt}

\noindent\begin{minipage}{\textwidth}
  \centering
  \captionsetup{skip=4pt}
  \includegraphics[width=0.9\linewidth]{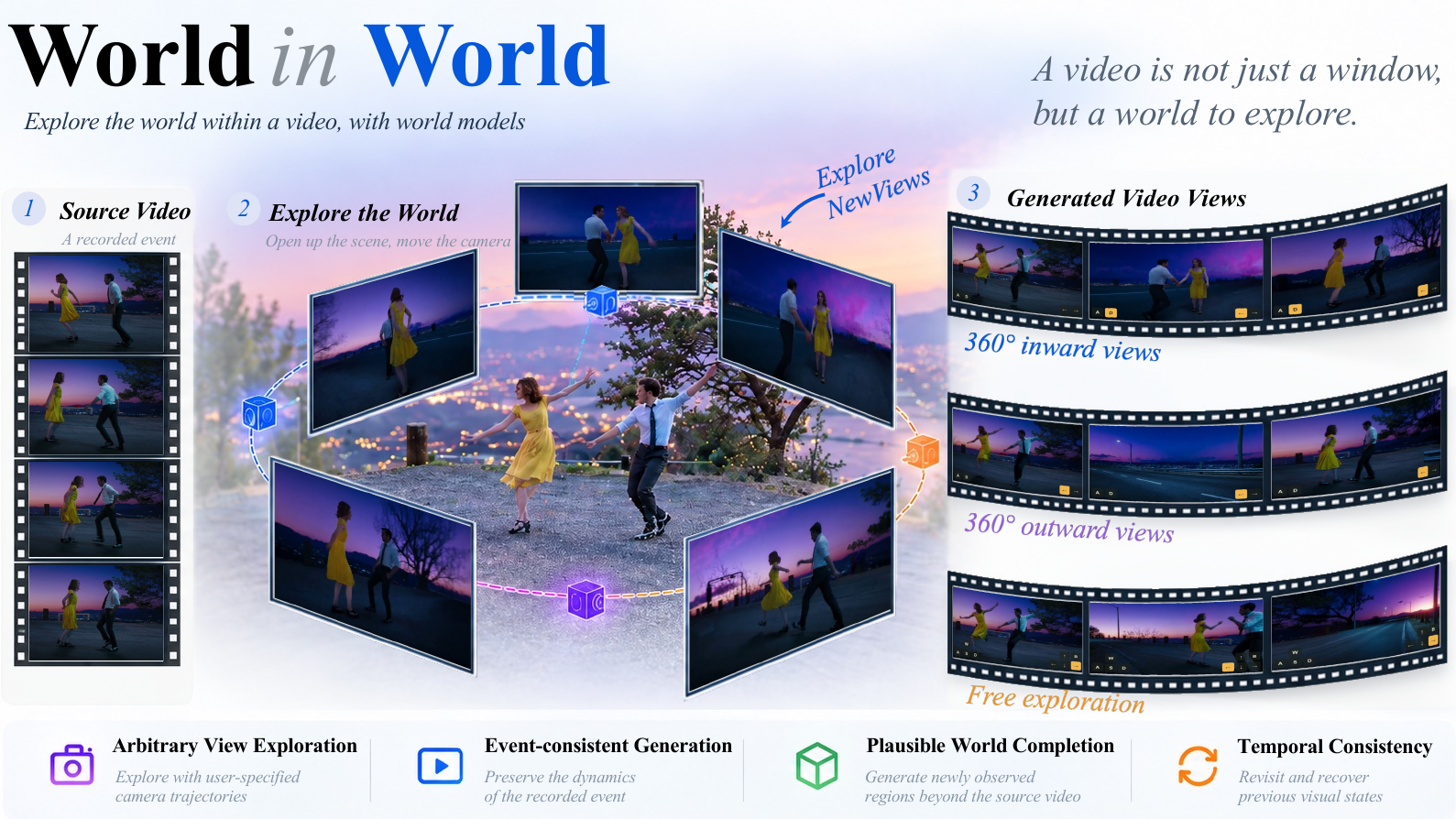}
  \captionof{figure}{%
    \textbf{\method{}: Explore the world within a video.}
    Video generation becomes an ongoing exploration of an evolving visual world.
    In \method{}, this exploration is grounded in the world of a given video,
    with the recorded event anchoring the appearance and dynamics of each new observation.
  }
  \label{fig:teaser}
\end{minipage}
\clearpage

\input{sec/introduction}
\input{sec/relatedwork}
\input{sec/methods}
\input{sec/experiments}
\input{sec/conclusion}

\clearpage
{
  \small
  \bibliographystyle{agilab}
  \bibliography{agilab}
}

\end{document}

%% file: math_commands.tex
\usepackage{amsmath,amsfonts,bm}

\def\eqref#1{equation~\ref{#1}}
\def\1{\bm{1}}

\DeclareMathAlphabet{\mathsfit}{\encodingdefault}{\sfdefault}{m}{sl}
\SetMathAlphabet{\mathsfit}{bold}{\encodingdefault}{\sfdefault}{bx}{n}

%% file: sec/abstract.tex
Autoregressive video world models enable interactive, long-horizon
exploration, but flexible control remains challenging. Exploring a source
video from new viewpoints requires the generated rollout to remain
synchronised with the recorded event, place observed content in the
requested view, plausibly complete newly exposed regions, and recover
previously generated appearance on revisits. Existing methods typically
address these requirements through task-specific modules or additional
training. We present \method{}, a training-free inference-time interface
that converts heterogeneous control evidence into camera- and time-labelled
clean visual states, which are read through the native self-attention
of a frozen causal video model. The evidence comprises source-video
observations, target-view scene projections, geometry renderings that
guide completion of newly exposed subject regions, and retrieved generated
states beyond the rolling cache. Each evidence source carries token-level
support and its own availability schedule. A correspondence router combines
persistent point identities with geometry to establish token
correspondences, guiding supported queries towards matching source-video
tokens. Evidence-wise attention CFG (EWA) then independently regulates each
auxiliary channel's additional contribution using attention responses
from the same denoising forward pass. The shared interface supports
camera-controlled rerendering, long-horizon revisiting, and human-motion
transfer with the same frozen backbone. We evaluate \method{} on
camera-controlled video rerendering under diverse viewpoint changes,
assessing perceptual quality, temporal consistency, and camera-following
accuracy.

%% file: sec/introduction.tex
\section{Introduction}
\label{sec:intro}

Recent advances in video generation have improved visual quality, motion
realism, and temporal coherence, and have supported the development of
video world models built on autoregressive generation
\citep{kong3D4DWorld2025,huangVid2WorldCraftingVideo2025,
lingbot-world,gaoInfiniteWorldsVersatile2026,
xuWonderVideoWorld2026,teamINSPATIOWORLDRealtime4D2026}.
While most text- or image-conditioned systems generate a finite clip from
conditions specified before inference, video world models support an ongoing
interactive process. They continually predict subsequent observations from
observed or generated visual states, allowing users to change camera
positions, viewing directions, and control actions throughout a rollout
\citep{sunWorldPlayLongTermGeometric2025,
teamAlayaWorldInteractiveLonghorizon2026}.
Video generation thus extends from producing a recording of an event to
supporting exploration of an evolving visual world, with applications in
interactive content creation, virtual production, game generation, and
embodied-agent simulation
\citep{li2025gamecraft,zhang2025matrixgame,he2025matrixgame2,
wang2026matrixgame3,qian2026matrixgame35}.
In particular, as large pretrained world models acquire stronger visual
and motion priors, a key question is how to flexibly extend their
controllability without retraining, enabling the same general world model
to accept diverse forms of visual control and world information and make
fuller use of its pretrained capabilities.

Supporting such diverse controls requires a world model to integrate
complementary visual evidence from observations, geometry, and generated
history, which differ in representation, spatial coverage, and temporal
relevance. As viewpoints and scene states change, generation must remain synchronised
with the dynamics specified by the visual evidence while maintaining correct
spatial placement and occlusion relationships in the target view.
Large viewpoint changes require using the model's generative prior to
complete unobserved scene and object surfaces, while long-horizon revisits
require recovering earlier appearance and spatial layout.
The challenge is therefore to make relevant evidence available and use it
at the appropriate locations and times throughout a rollout.

Existing approaches have explored camera conditioning, source-video
rerendering, geometric control, and long-term memory
\citep{zhou2026uniworldview,teamINSPATIOWORLDRealtime4D2026,
cameraanything2026,yu2025trajectorycrafter,bai2025recammaster,
xuWonderVideoWorld2026,xiaoWORLDMEMLongtermConsistent2025}.
Many rely on control-specific pathways or representations, so supporting
new evidence types or combinations often requires redesign or additional
training.
This motivates a shared interface through which the same pretrained world
model can use heterogeneous current and historical visual evidence without
learning a separate pathway for each control, which is the focus of our paper.

Our key observation is that causal
video world models equipped with a clean-state cache already have a
shared entry point for visual information: their native self-attention.
During autoregressive generation, the model reads the initial observation
and recently finalised outputs as clean visual states, while camera and
temporal encodings specify their viewpoints and temporal positions.
This suggests that external control information can be converted into
the same representation: clean visual states associated with camera poses,
event times, and valid spatial regions, made accessible through the
existing attention layers. Instead of adapting the model to each new
control, we express different controls in a visual representation that
the pretrained model already processes. Under this view, extending
world-model control becomes a visual evidence construction and
orchestration problem, rather than a model adaptation problem.

Based on this idea, we introduce \method{} (WiW), a
training-free visual-evidence interface for controlling frozen causal video
world models (Figure~\ref{fig:teaser}). WiW converts various visual
conditions, such as source observations, target-view projections, rendered
geometry, and generated history, into clean visual states annotated with
camera, temporal, and spatial-validity information. These states can be
directly accessed through the model's native self-attention, allowing
different sources of evidence to guide generation where and when they are
reliable. Our method is entirely training-free, requiring neither updates to
the pretrained world model nor learned control-specific modules. Under this
unified interface, flexible world-model control reduces to three
complementary problems: constructing useful visual evidence, localising the
relevant evidence, and regulating its influence on generation.

To provide the frozen world model with the complementary information required for controllable long-horizon rollouts, we first construct multiple forms of visual evidence for different control requirements. Our core idea is to let each evidence source contribute the information it can provide most reliably, while expressing all of them through the same clean-state interface that the pretrained model already understands.
Concretely, source-video observations provide appearance and content references from the recorded event. Since these observations do not explicitly determine where their content should appear under the target camera, \emph{target-view scene evidence} uses depth-based projection to place observed appearance in the requested view and provide spatial guidance within geometrically supported regions. When large camera motions expose subject surfaces that are absent from the source observations, \emph{rendered geometry evidence} supplies target-view shape and appearance proposals for the corresponding event state, helping the pretrained model complete newly visible subject surfaces. Finally, because a finite rolling cache eventually removes access to earlier generated states, we maintain a rollout-wide history and retrieve relevant and diverse archived states as \emph{generated evidence} when the camera returns. By assigning different control requirements to complementary evidence sources, WiW provides the frozen backbone with appearance references, spatial guidance, completion cues, and long-range visual context through a single shared visual interface.

To ensure that the model reads the right evidence and uses it with appropriate
strength, we further regulate how visual evidence participates in native
self-attention. Our core idea is to separately control \emph{where to read}
and \emph{how strongly to use} the available evidence, addressing errors in
evidence localisation and differences in evidence reliability.
For \emph{where to read}, appearance-based attention can become ambiguous
under large viewpoint changes, dynamic motion, and repeated textures.
We therefore introduce \emph{correspondence-guided attention routing}
(CGAR), which uses persistent point identities and camera geometry to route
current queries towards geometrically corresponding source-video tokens when
valid matches are available. For \emph{how strongly to use}, different
evidence sources can vary in reliability across spatial regions and denoising
stages. We introduce \emph{evidence-wise attention CFG} (EWA), which compares
native and evidence-conditioned attention responses to independently amplify
compatible and complementary information while suppressing excessive
guidance. Both mechanisms operate directly within the model's native
self-attention, and EWA reuses responses from the same denoising forward pass,
introducing no additional network function evaluations (NFE) for guidance.

Through this formulation, WiW provides a unified, training-free framework
for extending the control capabilities of frozen world models through visual
evidence, without introducing control-specific training or adaptation.
Notably, beyond exploring worlds freely generated by a world model, WiW
enables exploration \emph{within the world of a given video}: users can
navigate the recorded dynamic world along new camera trajectories while
preserving the appearance and temporal progression of the original event.
We instantiate WiW on the publicly released causal-fast checkpoint of
LingBot-World~2.0~\citep{gaoInfiniteWorldsVersatile2026}, with all pretrained
parameters frozen, and evaluate camera-controlled video rerendering on DAVIS
and OpenVid-1M~\citep{ponttuset2017davis,nan2024openvid} across diverse
viewpoint changes. Beyond rerendering, WiW supports applications including
bullet-time generation, video stabilization, video editing, and K/V sharing between two generation cases produced by the same frozen model, demonstrating the versatility of the proposed interface.

Our main contributions are:
\begin{itemize}
    \item We introduce WiW, a training-free visual-evidence interface for flexibly extending the control capabilities of frozen causal video world models.

    \item We develop complementary visual evidence that enables event-synchronised, spatially aligned, and long-horizon-consistent exploration of dynamic video worlds.

    \item We introduce correspondence-guided attention routing (CGAR) and evidence-wise attention CFG (EWA) to localise and regulate heterogeneous visual evidence within native self-attention.

    \item We demonstrate flexible exploration of given video worlds across diverse camera trajectories and downstream applications using a single frozen world-model backbone.
\end{itemize}

%% file: sec/relatedwork.tex
\section{Related work}
\label{sec}

\subsection{World models}

A video world model predicts the visual observations an observer would
receive while moving through an environment, given an initial observation
and a requested camera or action. In static environments, a key requirement
is to maintain consistent scene structure and appearance as the viewpoint
changes and previously seen regions are revisited. One line of work uses
explicit scene representations, such as point clouds, meshes, or Gaussians,
to organize observations in a shared coordinate system and render them
from the target camera. These representations provide a clear spatial
reference, but require completing unobserved regions and maintaining the
scene state as exploration continues
\citep{kong3D4DWorld2025,yuWonderJourneyGoingAnywhere2024,yuWonderWorldInteractive3D2024,liangWonderlandNavigating3D2024,wangVistaDreamSamplingMultiview2024,yuViewCrafterTamingVideo2024,chenFlexWorldProgressivelyExpanding2025,huangTerraExplorableNative2025,renGEN3C3DinformedWorldconsistent2025,bahmaniLyraGenerative3D2025,huangGen3R3DScene2026,gaoPixWorldUnifying3D2026,liFlashWorldHighquality3D2025,hy-worldHYworld20Multimodal2026}.
Another line of work directly predicts target-view observations with video
models, using learned generative priors to handle viewpoint changes and
newly visible regions
\citep{sunDimensionXCreateAny2024,zhaoGenXDGeneratingAny2024,wangVideoSceneDistillingVideo2025,zhuGTAAdvancingImageto3D2026,holleinWorldReconstructionInconsistent2026}.
When the environment changes over time, the model must also maintain
consistency between viewpoint changes, scene structure, and event
progression. In dynamic environments, methods need to generate observations
that match both the target viewpoint and the current state of the recorded
or simulated event. Previous work has studied this problem through dynamic
scene generation, source-video re-observation, and continuous world modeling
\citep{chen4DNeXFeedforward4D2025,daiFantasyWorldGeometryconsistentWorld2025,xiangVideoWeaveUnlockingGeometric2026,zhengVerseCrafterDynamicRealistic2026,bai2025recammaster,teamINSPATIOWORLDRealtime4D2026,paliwalReshootanythingSelfsupervisedModel2026,xuRealCamRealtimeNovelview2026,linVista4DVideoReshooting2026,huEX4DEXtremeViewpoint2025,luSEE4DPosefree4D2025}.
Autoregressive and self-forcing methods further extend scene generation
to continuous rollouts, requiring models to maintain consistent appearance,
spatial relationships, and event states over long explorations
\citep{chenDiffusionForcingNexttoken2024,huangSelfForcingBridging2025,huangVid2WorldCraftingVideo2025,maoYumeInteractiveWorld2025,chenDeepVerse4DAutoregressive2025,teamAlayaWorldInteractiveLonghorizon2026,gaoInfiniteWorldsVersatile2026}.
Pretrained video models have appearance and motion priors that generalize
to new scenes, but their native conditioning interfaces are usually
determined by the inputs used during training. The visual-history pathway
can itself serve as a control interface:
Warp-as-History~\citep{wangWarpashistoryGeneralizableCameracontrolled2026}
feeds camera-warped observations as pseudo-history, aligns their temporal
positions with target frames. We build on a pretrained causal video model and study how
to reuse its native self-attention mechanism for reading visual states,
allowing it to accept additional control information while keeping its
parameters frozen.

\subsection{Conditioning mechanisms for video world models}

Existing video world models usually use dedicated conditioning mechanisms to introduce different types of control into generation. For camera control, prior methods represent camera trajectories as rays, poses, positional encodings, projected features, or rendered geometric proxies, and use corresponding conditioning modules to guide generation \citep{jangRaysPixelsLearning2026,liReRoPERepurposingRoPE2026,xiangGeometryawareRotaryPosition2026,lee3DScenePrompting2025,liuCamGeoSparseCameraconditioned2026,wuGeometryForcingMarrying2025,guDiffusionShader3Daware2025,kimVideoFrom3D3DScene2025}. Source-video rerendering extends control from camera motion to re-observation of dynamic content. These methods often train additional modules for source-video inputs to bring the original event's appearance and dynamics into the target view \citep{bai2025recammaster,chenPostCamCameracontrollableNovelview2025,seoVidCamEditVideoCamera2025,teamINSPATIOWORLDRealtime4D2026,wang2026directing}. Similarly, the concurrent work Wonder supports image- and video-conditioned world generation by jointly training a rendered control field, a sparse memory, and a distilled causal student to combine multiple conditions \citep{xuWonderVideoWorld2026}. These methods show that dedicated conditioning pathways can support specific control tasks, but different types of control often use different input interfaces and training procedures. Beyond current observations and external controls, continuous exploration also requires access to scene information generated earlier. Since causal models typically read only a limited recent context, prior work uses explicit geometric memory, persistent states, or historical information retrieval to recover scene content beyond the current context and constrain subsequent generation \citep{liVMemConsistentInteractive2025,wangEvoWorldEvolvingPanoramic2025,weiGeometryawareImplicitMemory2026a,huangMemoryForcingSpatiotemporal2025,xiaoWORLDMEMLongtermConsistent2025,yuContextMemorySceneconsistent2025,jooRetrieveWhatsMissing2026,yiWorldKVEfficientWorld2026a}. Memory therefore helps maintain long-term consistency and can also serve as a visual condition alongside current observations and geometry. WiW follows this idea by including historical information in a unified visual conditioning framework, allowing scene evidence from different sources to jointly guide subsequent generation.

%% file: sec/methods.tex
\begin{figure*}[t]
    \centering
    \includegraphics[width=\textwidth]{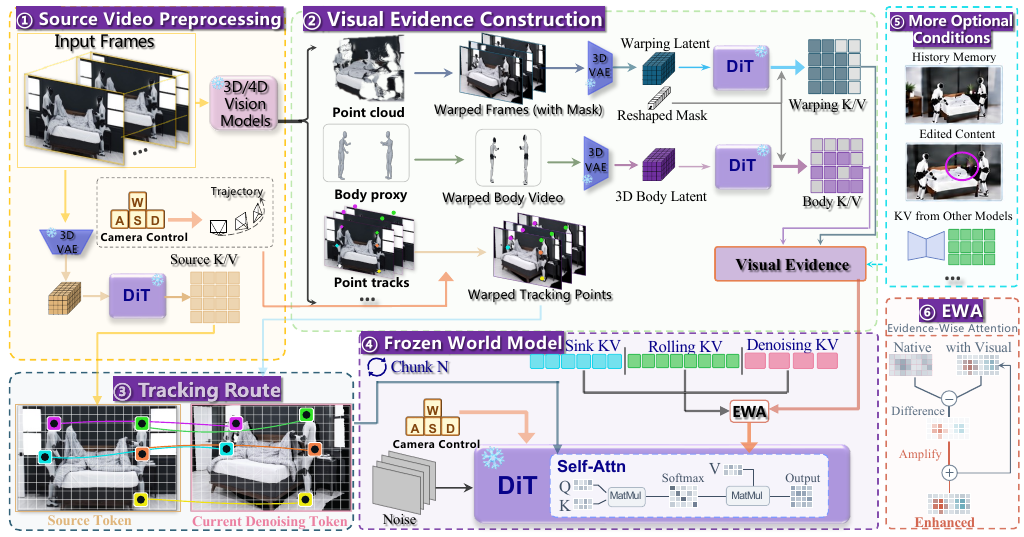}
    \caption{
        \textbf{Overview of the \method{} pipeline.}
        Target-view
        projections, rendered geometry, and retrieved historical states
        provide temporary visual evidence. Correspondence-guided attention
        routing connects queries to matching source-video tokens, and
        evidence-wise attention CFG regulates each channel's additional
        contribution. Temporary evidence blocks are removed after each
        chunk; finalized outputs enter the rolling cache, and their visual
        features are archived after eviction.
    }
    \label{fig:ww_overview}
\end{figure*}

\section{Method}
\label{sec:method}

We propose \method{} (WiW), a unified visual-evidence interface for
extending the control capabilities of frozen causal video world models.
We illustrate the framework through camera-controlled video rerendering:
given a source video of a dynamic event and a target camera trajectory $C$,
we aim to rerender the event from the requested viewpoints while preserving
its appearance and temporal progression. To this end, we convert source
observations, geometry, and generated history into clean visual states
with camera, temporal, and spatial-validity information. The frozen model
can then read this evidence through native self-attention, without
additional training or learned control-specific adapters.

Figure~\ref{fig:ww_overview} summarizes the overall pipeline. We first define the shared evidence interface and explain how visual evidence participates in native self-attention (Section~\ref{sec:control_as_context}). Building on this
interface, we project source observations into the target view to provide
layout references (Section~\ref{sec:target_view_context}), use rendered
geometry to guide completion of newly exposed subject surfaces
(Section~\ref{sec:rendered_geometry_context}), and retrieve generated
history for consistent long-horizon revisits
(Section~\ref{sec:retrieved_context}). Together, these sources provide
complementary references for generation. Attention routing and evidence-wise
attention CFG then localize relevant evidence and regulate its influence,
respectively (Section~\ref{sec:correspondence_routing}).

\subsection{Control through Visual Evidence}
\label{sec:control_as_context}

When generating the current chunk, our causal video backbone reads the
initial observation and recently finalized states through native
self-attention. We use this existing pathway to convert multiple forms
of visual evidence, including source-video observations, target-view
scene projections, rendered geometry, and generated history, into the same
type of states that the model already reads. These sources differ in
representation, valid spatial extent, and temporal coverage. Their shared
representation must therefore specify evidence content, camera and temporal
information, spatial support, and channel activation across denoising stages.

To represent this information consistently, at denoising step $s$, converter
$\Psi_b$ maps the raw evidence $E_b$ of channel $b$ and target camera
trajectory $C$ to:
\begin{equation}
    \Psi_b(E_b;C,s)
    =
    \left(
        X_b,\,
        P_b,\,
        \boldsymbol{\omega}_b,\,
        \gamma_b^{(s)}
    \right).
    \label{eq:ww_context_interface}
\end{equation}
Here, $X_b$ denotes the visual content of the evidence; $P_b$ records
per-frame camera intrinsics, poses, and event-time indices;
$\omega_{bj}\in[0,1]$ specifies the spatial support of evidence token
$j$; and $\gamma_b^{(s)}\in\{0,1\}$ determines whether the channel is active
at step $s$.

To turn this representation into attention features that the model can
read, we feed the clean visual states corresponding to $X_b$, together
with camera and temporal information $P_b$, into the frozen video backbone
with the diffusion timestep set to $t=0$. A single network forward pass
extracts the keys $K_b^\ell$ and values $V_b^\ell$ from each attention layer,
which we cache as \emph{clean K/V}. During target-chunk generation, queries
from the current denoising states read these cached features through
self-attention, allowing external visual evidence to guide generation.
All evidence types introduced below use this procedure to obtain their
attention features.

These K/V features participate in attention through either the native
block or temporary auxiliary blocks. The native block retains the
backbone's 18 latent-frame-state slots: six source anchors, eight
recent-history states, and four states in the current chunk. Source-video
evidence occupies the source-anchor slots, while other evidence is attached
as temporary auxiliary attention blocks and removed once the current chunk
is finalized. All blocks reuse the frozen backbone's query/key/value
(Q/K/V) projections, while preserving its native text- and
camera-conditioning pathways.

When visual evidence enters attention, its temporal relationship to the current generation states needs to be specified. We therefore set the temporal coordinates of rotary position embeddings (RoPE) and apply the corresponding rotations to attention queries and keys, incorporating relative temporal information into attention.

Spatial support weights regulate the contribution of evidence tokens by
scaling their unnormalized attention weights by $\omega_{bj}$.
Channel activation $\gamma_b^{(s)}$ further determines the denoising
stages in which each evidence source participates.

\subsection{Target-View Scene Evidence}
\label{sec:target_view_context}

Source-video observations provide appearance references for the recorded event, but they remain in the source-camera view and do not explicitly determine where their content should appear in the target image. Camera motion changes projected positions and occlusion relationships, while ambiguous cross-view appearance matches can cause visible structures to drift. To provide an explicit spatial layout reference, we project source observations into the target view, allowing the model to read observed appearance aligned with the requested camera.

To construct this reference for target frame $m$ and temporally aligned
source view $r$, we use source RGB $I_{r,m}^{\mathrm{src}}$ and estimate
its depth $D_{r,m}^{\mathrm{src}}$ with
DepthCrafter~\citep{hu2025depthcrafter}. We back-project source pixels into
3D and reproject them from source camera $C_{r,m}^{\mathrm{src}}$ to target
camera $C_m=(A_m,T_m)$, where $A_m$ denotes camera intrinsics and $T_m$ is
the absolute camera-to-world pose:
\begin{equation}
    \left(
        X_{r,m}^{\mathrm{proj}},
        M_{r,m}^{\mathrm{proj}},
        Z_{r,m}^{\mathrm{proj}}
    \right)
    =
    \operatorname{Proj}\!\left(
        I_{r,m}^{\mathrm{src}},
        D_{r,m}^{\mathrm{src}};
        C_{r,m}^{\mathrm{src}}\rightarrow C_m
    \right).
    \label{eq:ww_target_projection}
\end{equation}
Projection uses a shared coordinate system and scale. The outputs
$X_{r,m}^{\mathrm{proj}}$, $M_{r,m}^{\mathrm{proj}}$, and
$Z_{r,m}^{\mathrm{proj}}$ are target-view RGB, a binary visibility mask,
and target-camera depth, respectively. Following the shared interface,
we encode the projected RGB with target-camera and event-time information,
and use visibility to determine the evidence's spatial support.

Since occlusion and local geometric errors can affect the projection,
we determine its spatial support from visibility and geometric reliability,
then convert it into token-level support weights
$\boldsymbol{\omega}_{\mathrm{proj}}$. These weights allow the model to use
reliable target-view layout references while reducing the influence of
uncertain regions.

\subsection{Rendered Geometry Evidence}
\label{sec:rendered_geometry_context}

Target-view projection reorganizes content already present in the source
observations, but its coverage remains limited by source-view visibility.
When camera motion exposes subject surfaces not observed in the source video,
these regions lack direct shape and appearance references, and their
generation may deviate from the subject's structure or current pose.
To provide evidence for completing such regions, we use a renderable
subject representation to produce shape and appearance proposals from the
target camera at the corresponding event time.
Given subject state $G_m$ at event time $m$ and target camera $C_m$,
rendering yields:
\begin{equation}
    \left(
        X_m^{\mathrm{geo}},
        M_m^{\mathrm{geo}},
        Z_m^{\mathrm{geo}}
    \right)
    =
    \operatorname{Render}(G_m,C_m).
    \label{eq:ww_geometry_render}
\end{equation}
Here, $X_m^{\mathrm{geo}}$, $M_m^{\mathrm{geo}}$, and $Z_m^{\mathrm{geo}}$
denote RGB, a geometric support mask, and target-camera depth, respectively.
The rendering is encoded through the shared interface, and its support
mask is converted into token-level support weights.

For example, when the subject is human, we represent its state as
$G_m=(\mathcal A,\Theta_m)$. Here, $\mathcal A$ is an avatar reconstructed
with LHM++~\citep{qiu2025lhmpp} from sharp, minimally occluded full-body
source-video crops, and $\Theta_m$ contains the per-frame
SMPL-X~\citep{pavlakos2019smplx} parameters that drive the avatar.
We align the avatar and body model to the projected scene's coordinate
frame and depth scale using depth correspondences in the source view.
LHM++ provides target-view RGB and support masks for observed and inferred
unseen surfaces, while the aligned SMPL-X geometry provides depth for
resolving occlusion between subjects. We also compare this depth with
the projected scene depth to remove unreliable background projections
near subject boundaries.

\subsection{Retrieved Generated Evidence}
\label{sec:retrieved_context}

Scene projections and geometry renderings provide observed and
reconstruction-based evidence for the current target view. As generation
progresses, previously generated appearance and layout also become
important references for subsequent frames. The native rolling cache
retains only recent states, so states corresponding to an earlier region
may have been evicted when the camera returns, leaving their generated
content inaccessible to the model. To recover these references, we
maintain a rollout-wide history bank and retrieve finalized states relevant
to the current view.

The history bank accumulates evidence as the rolling cache is updated.
For each evicted finalized state, we archive its per-layer clean K/V
with the associated camera pose and temporal index. Since the bank
grows throughout generation, we select a bounded number of relevant states
to participate in attention for each target chunk. Specifically, we rank
historical states by target-view surface coverage and viewing-direction
compatibility, then select a diverse top-$K_{\mathrm{ret}}$ subset to reduce
redundant references. The model reads the retrieved K/V directly
through temporary auxiliary attention blocks, recovering historical
appearance and layout beyond the rolling cache to maintain scene
consistency.

\subsection{Attention Routing and Guidance}
\label{sec:correspondence_routing}
\label{sec:attention_cfg}

The construction and retrieval steps above provide complementary evidence
for the source event, target-view layout, subject geometry, and generated
history. However, making evidence accessible does not ensure that queries
locate the relevant positions within it. Viewpoint changes and dynamic
motion alter the appearance of a surface, while repeated textures can make
distinct locations look similar, introducing ambiguity into
appearance-based attention. We therefore first use geometric
correspondences to match current generation tokens with source-video
tokens and guide evidence reading, then regulate each channel's additional
influence through its attention response.

\noindent \textbf{Correspondence-guided attention routing.} To associate
queries from current generation tokens that have valid geometric
correspondences with matching source-video locations, we introduce
correspondence-guided attention routing (CGAR). We track points with
persistent identities in the source video and combine depth and camera
information to establish token-level correspondences between the current
view and the source video. For a geometrically matched query $i$ and
source-video key $j$, we add the logarithm of correspondence weight
$\beta_{ij}>0$ to the routing attention score:
\begin{equation}
    \ell_{ij}^{\mathrm{corr}}
    =
    \frac{\mathbf q_i^{\top}\mathbf k_j}{\sqrt{d_h}}
    +\log\beta_{ij}.
    \label{eq:ww_routing_logit}
\end{equation}
Here, $d_h$ is the attention-head dimension, and $\beta_{ij}>0$ controls
the contribution of the correspondence. We combine the routing response
with the other attention-block responses through joint normalization,
allowing current generation queries to read geometrically matched
source-video evidence.

\noindent \textbf{Evidence-wise attention CFG.} The evidence sources
constructed above participate in video generation through self-attention
during denoising. Different auxiliary evidence channels provide
constraints on target-view layout, subject structure, and generated
history, and the information they provide also differs in reliability
and importance. Their influence on generation therefore needs to be
controlled independently. Using all evidence with the same strength
makes it difficult to balance adherence to different conditions with
the model's generative prior. Inspired by classifier-free guidance
(CFG)~\citep{ho2022classifierfree} and NAG~\citep{chen2026normalized}, we
propose evidence-wise attention CFG (EWA). EWA independently adjusts the
guidance strength of each evidence source at the level of attention
responses, strengthening useful control signals while preserving the
content quality and naturalness of native generation.

Specifically, at the same attention layer and for the same queries, let
$\mathbf o_0$ denote the response obtained by attending only to the native
block, and let $\mathbf o_e$ denote the jointly normalized response obtained
by attending to both the native block and evidence $e$. Directly amplifying
$\mathbf o_e$ would also amplify its component along the native generation
direction. We therefore strengthen only the complementary direction
introduced by the evidence relative to the native response. We first
remove the projection of $\mathbf o_e$ onto $\mathbf o_0$, modulate the
remaining component by the nonnegative cosine similarity between the two
responses, and use an independent guidance strength $g_e$ to control the
correction magnitude. The updated response for evidence $e$ is:
\begin{equation}
    \widetilde{\mathbf o}_e
    =
    \mathbf o_e
    +
    g_e\,\max\!\left(0,\cos(\mathbf o_e,\mathbf o_0)\right)
    \left[
        \mathbf o_e
        -\operatorname{Proj}_{\mathbf o_0}(\mathbf o_e)
    \right].
    \label{eq:ww_attention_cfg}
\end{equation}
Here, $\cos(\mathbf o_e,\mathbf o_0)$ is the cosine similarity
between the two responses. Removing the projection along the native
response direction allows EWA to strengthen the complementary information
provided by the evidence without repeatedly amplifying the model's
existing response. 

We then add the evidence-specific corrections to the joint response of all
active attention blocks and bound the output magnitude using the norm of
the native response $\mathbf o_0$ as a reference, preventing excessive
guidance from disrupting generation.

EWA operates directly on attention responses within the same denoising
forward pass. Its computation reuses existing attention-block outputs and
normalization results, without requiring a separate full denoising-network
forward pass for each evidence source. It therefore adds no network
function evaluations (NFE) for guidance.

%% file: sec/experiments.tex
\section{Experiments}
\label{sec:experiments}

\providecommand{\figplaceholder}[2]{%
  \fbox{\parbox[c][#1][c]{0.96\linewidth}{\centering #2}}}

We evaluate \method{} through camera-controlled video rerendering, where
a source video serves as visual evidence for exploring the recorded world
along a target camera trajectory. We further examine long-horizon
revisiting and human-motion transfer, and ablate individual components
to assess how the same frozen backbone uses complementary visual evidence.

\subsection{Experimental Setup}
\label{sec:setup}

\noindent\textbf{Implementation.}
We instantiate \method{} on the publicly released causal-fast checkpoint
of LingBot-World~2.0~\citep{gaoInfiniteWorldsVersatile2026}, keeping all
pretrained parameters frozen. Visual evidence is constructed and attached
at inference time, without additional training or learned control-specific
adapters.

\noindent\textbf{Baselines.}
Following prior work on camera-controlled video rerendering, we select videos from
DAVIS~\citep{ponttuset2017davis} and OpenVid-1M~\citep{nan2024openvid}
and compare with ReCamMaster, TrajectoryCrafter, WorldForge,
InSpatio-World, UniWorld-View, and CameraAnything
\citep{bai2025recammaster,yu2025trajectorycrafter,song2026worldforge,
teamINSPATIOWORLDRealtime4D2026,zhou2026uniworldview,cameraanything2026}.
Each baseline uses its official implementation, released checkpoint,
and recommended configuration. ReCamMaster and TrajectoryCrafter generate
81 and 49 frames, respectively, as required by their official implementations;
the remaining methods use their officially recommended frame counts.
All methods receive the same source video and continuous target camera
path, with camera parameters converted to the convention required by each
method. Before evaluation, we align the generated videos across methods
in terms of camera angles and frame count. All methods share the same
metric preprocessing pipeline.

\noindent\textbf{Metrics.}
We evaluate image fidelity using PSNR, SSIM, and LPIPS.
To assess video quality, temporal consistency, and dynamics,
we use the official VBench implementation~\citep{huang2023vbench}
to report seven dimensions: Aesthetic Quality, Imaging Quality,
Temporal Flickering, Motion Smoothness, Subject Consistency,
Background Consistency, and Dynamic Degree.
We additionally report \emph{Overall},
computed as the average score of these seven dimensions.

Following InSpatio-World~\citep{teamINSPATIOWORLDRealtime4D2026},
we evaluate camera control using translation error (TransError)
and rotation error (RotError).
We independently estimate camera trajectories from the generated videos
using Depth Anything~3 and ViPE~\citep{lin2025depthanything3,huang2025vipe}.
For each estimator, we compare the estimated trajectories with
the prescribed target trajectories.
The reported TransError and RotError are obtained by averaging
the corresponding errors across the two estimators.

\subsection{Quantitative Comparisons}
\label{sec:quantitative}

Table~\ref{tab:rerendering_quant} reports quantitative results for
camera-controlled video rerendering on DAVIS and OpenVid-1M.
Following previous state-of-the-art, we evaluate performance in terms of VBench, camera trajectory errors and image quality.

\begin{table*}[t]
  \centering
  \caption{Quantitative comparison of camera-controlled video rerendering
  on DAVIS and OpenVid-1M, evaluated using VBench dimensions,
  camera trajectory errors, and image fidelity metrics.
  Overall is the average score of all seven reported VBench dimensions.
  \textbf{Bold} and \underline{underline} indicate the best and
  second-best results, respectively. Tied results share the same rank.}
  \label{tab:rerendering_quant}
  \small
  \setlength{\tabcolsep}{2.0pt}
  \renewcommand{\arraystretch}{1.08}
  \resizebox{\linewidth}{!}{%
    \begin{tabular}{lccccccccccccc}
      \toprule
      & \multicolumn{8}{c}{VBench $\uparrow$}
      & \multicolumn{2}{c}{Camera errors $\downarrow$}
      & \multicolumn{3}{c}{Image quality} \\
      \cmidrule(lr){2-9}
      \cmidrule(lr){10-11}
      \cmidrule(lr){12-14}
      Method
      & Aesth. & Img. & Flick. & Smooth. & Subj. & Bg.
      & Dyn. & Overall
      & TransError & RotError ($^\circ$)
      & PSNR $\uparrow$ & SSIM $\uparrow$ & LPIPS $\downarrow$ \\
      \midrule

      ReCamMaster
      & 54.283 & 66.290 & \textbf{95.178} & \underline{97.503}
      & 88.941 & \underline{92.154}
      & 92.500 & 83.836
      & 0.124289 & 7.8626
      & 15.1383 & 0.444346 & 0.377962 \\

      TrajectoryCrafter
      & 51.322 & 61.466 & 94.169 & 97.423 & 87.032 & 90.727
      & \textbf{98.750} & 82.984
      & 0.090812 & 5.3254
      & 19.9689 & 0.625112 & 0.217875 \\

      WorldForge
      & 49.915 & 62.356 & 94.262 & 96.934 & 84.683 & 90.017
      & \underline{98.524} & 82.384
      & 0.079961 & 4.5021
      & 19.6494 & 0.589250 & 0.218029 \\

      InSpatio-World
      & 53.162 & 66.076 & 93.622 & 96.764 & 87.070 & 91.188
      & 96.250 & 83.447
      & 0.082399 & 4.4011
      & 20.6855 & 0.613398 & 0.183115 \\

      UniWorld-View
      & 54.237 & \underline{67.534} & 93.444 & 96.627
      & \underline{88.946} & 91.777
      & 97.500 & \underline{84.295}
      & \underline{0.068705} & 4.1958
      & \underline{22.4735} & \underline{0.770554}
      & \underline{0.121668} \\

      CameraAnything
      & \underline{54.365} & 67.526 & \underline{94.489}
      & 96.908 & 87.399 & 91.817
      & 88.750 & 83.036
      & 0.186443 & \underline{3.1868}
      & 15.1583 & 0.453712 & 0.342180 \\

      \textbf{Ours}
      & \textbf{56.556} & \textbf{68.004} & 93.711
      & \textbf{97.749} & \textbf{88.948} & \textbf{92.627}
      & \textbf{98.750} & \textbf{85.192}
      & \textbf{0.068622} & \textbf{2.8326}
      & \textbf{23.1511} & \textbf{0.787205}
      & \textbf{0.121664} \\

      \bottomrule
    \end{tabular}%
  }
\end{table*}

\subsection{Qualitative Comparisons}
\label{sec:qualitative}

\begin{figure*}[t]
  \centering
  \includegraphics[width=\linewidth]{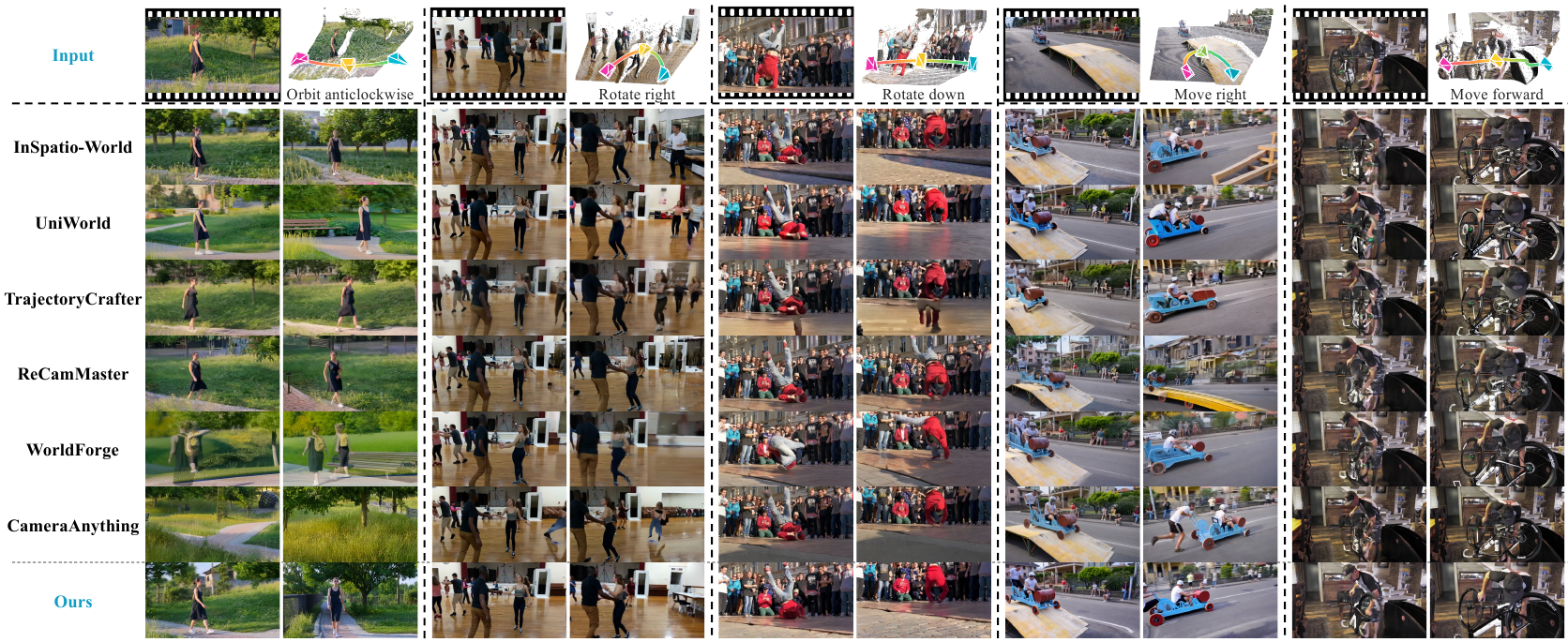}
  \caption{Qualitative comparisons under diverse camera motion. All methods receive the same source video and
  target camera path. Representative frames illustrate subject and scene
  consistency, viewpoint control, and completion of newly exposed regions.}
  \label{fig:qual_video}
\end{figure*}

Figure~\ref{fig:qual_video} compares camera-controlled rerendering under different camera motion. Each example presents
the source video and representative outputs under the same target camera path. 

\subsection{Additional Applications}
\label{sec:extensions}

\begin{figure*}[t]
  \centering
  \includegraphics[width=\linewidth]{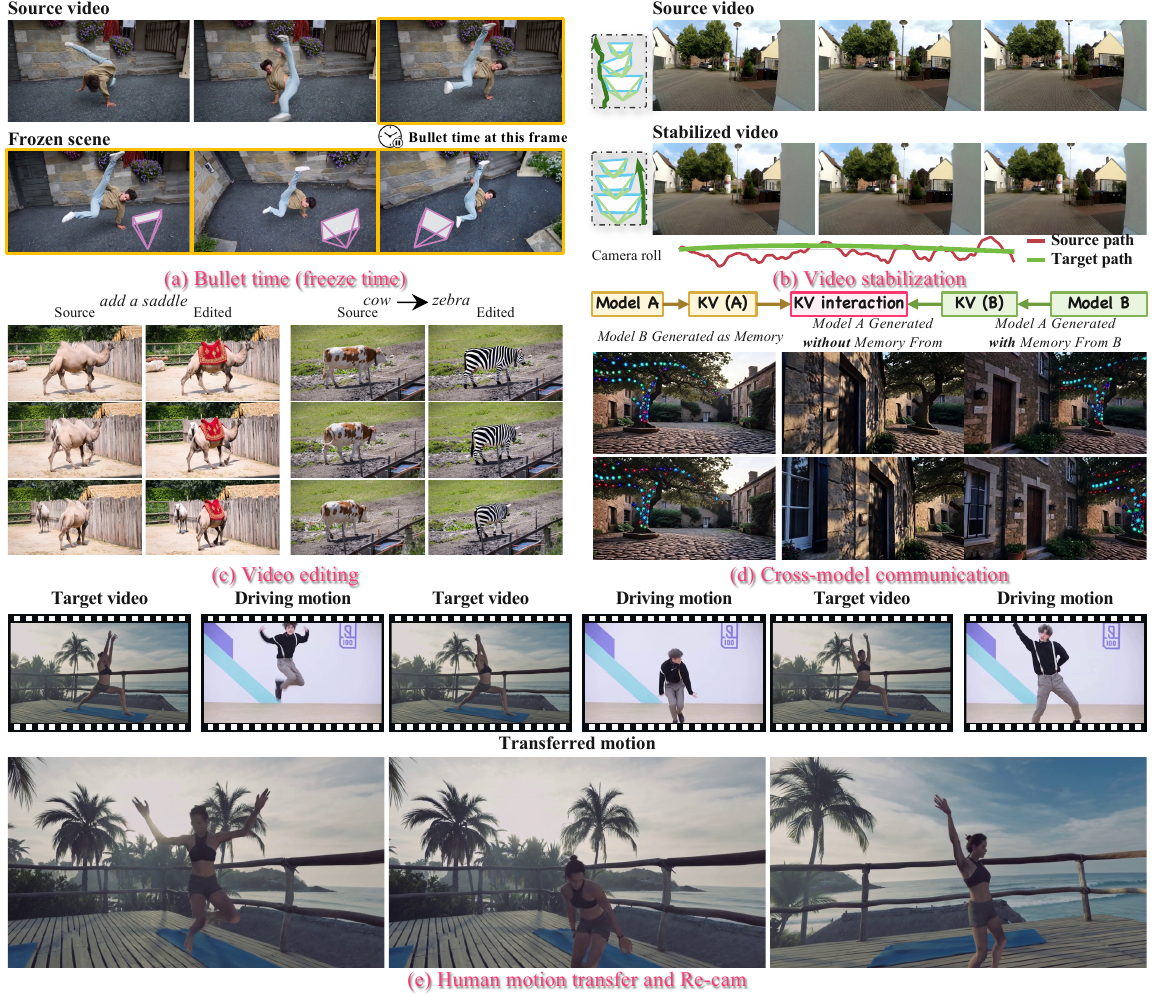}
  \caption{Applications of the same visual-evidence interface, including
  bullet-time rendering, video stabilization, video editing, K/V sharing
  between generated cases and human motion transfer. Model A and Model B denote two instances of
  the same frozen model running different generation cases. The instances
  communicate by sharing cached K/V as visual evidence.}
  \label{fig:extensions}
\end{figure*}

Our framework expresses different control requirements through a shared
visual-evidence interface.
Figure~\ref{fig:extensions} extends this idea beyond rerendering to
control over event timing, changes to appearance and motion, and the
sharing of generated visual information. In the K/V-sharing example,
Model A and Model B denote two instances of the same frozen model, each
corresponding to a different generation case. Communication between the
instances is implemented by passing cached K/V from one instance to the
other as visual evidence to guide generation.
Across these applications, different sources of evidence guide generation
through the same native attention mechanism, with camera and time
information specifying how that evidence relates to the requested output.
This allows us to extend the model's control capabilities while keeping
the backbone frozen, without task-specific training.

\subsection{Ablation Studies}
\label{sec:ablation}

\begin{figure*}[t]
  \centering
  \includegraphics[width=\linewidth]{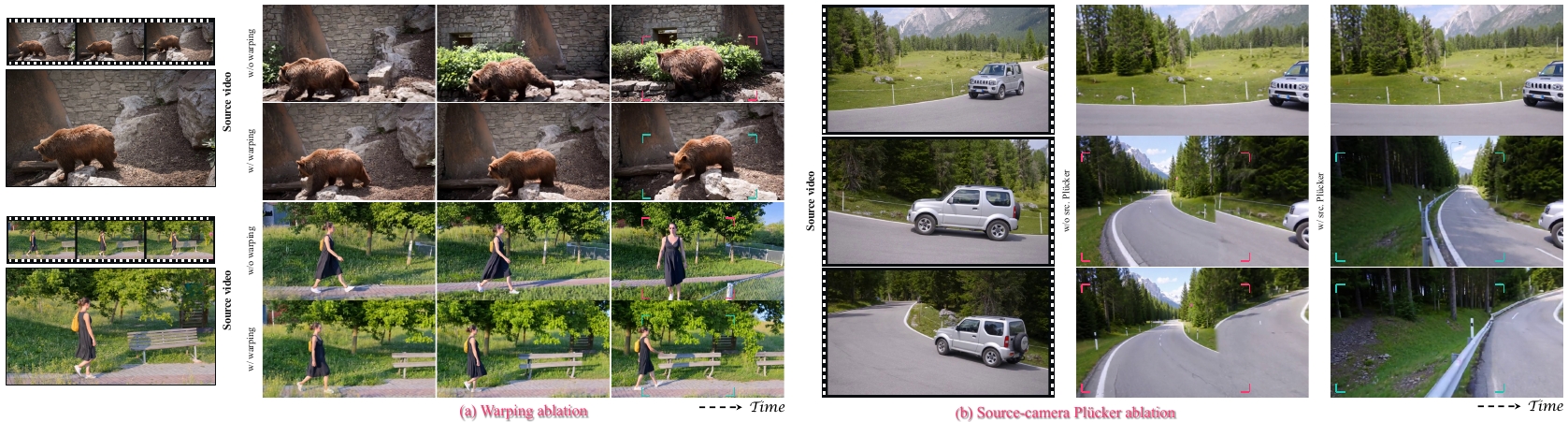}
  \caption{Qualitative ablations of spatial evidence and camera conditioning.
  \textbf{(a)} Target-view warping helps preserve subject structure
  and scene layout under viewpoint changes.
  \textbf{(b)} Source-camera Pl\"ucker conditioning helps maintain
  cross-view structure by retaining source-view camera information.}
  \label{fig:ablation_warping_plucker}
\end{figure*}

\begin{figure*}[t]
  \centering
  \includegraphics[width=\linewidth]{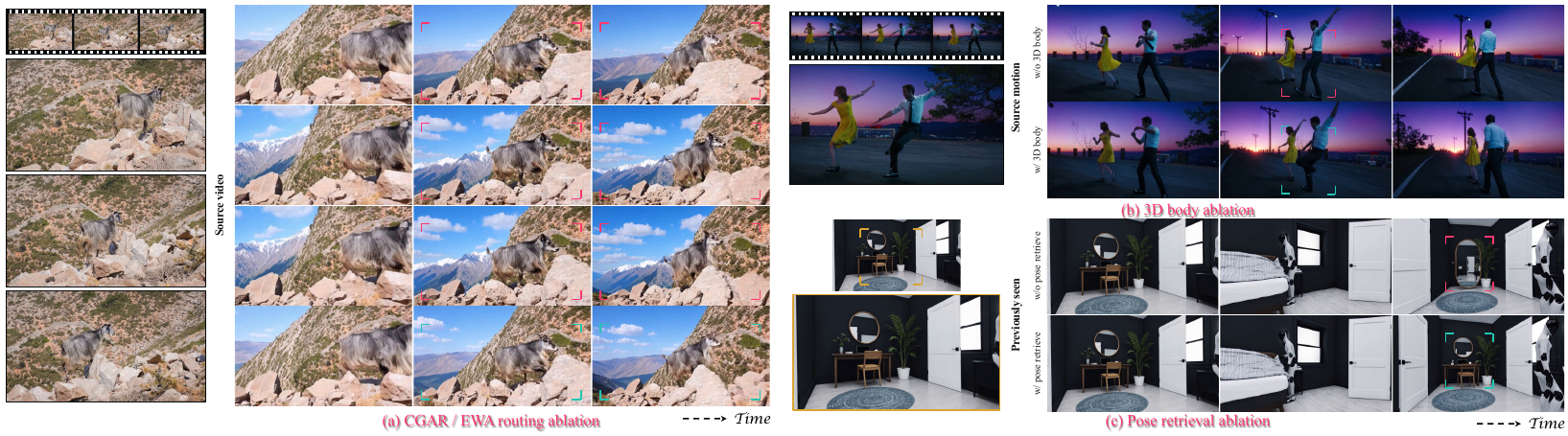}
  \caption{Qualitative ablations of evidence use and availability.
  \textbf{(a)} Correspondence-guided attention routing (CGAR) and
  evidence-wise attention (EWA) help preserve subject appearance
  and background structure.
  \textbf{(b)} Rendered body geometry guides the completion of newly
  exposed subject regions.
  \textbf{(c)} The camera revisits a previously generated region after the
  earlier states have been removed from the rolling cache. Historical
  retrieval helps preserve the appearance and layout established during
  the earlier visit.}
  \label{fig:ablation_evidence_usage}
\end{figure*}

We examine how the choice and use of visual evidence contribute to
controllable generation.
Table~\ref{tab:quantitative_ablation} shows that the full method achieves
the lowest camera errors and the best or tied-best results on all
reported VBench dimensions. Accurate camera control requires a clear spatial reference for the
observed content.
Figure~\ref{fig:ablation_warping_plucker} examines target-view warping
and source-camera Pl\"ucker conditioning from this perspective.
Removing warping causes the largest degradation across all reported
metrics, with rotation and translation errors rising to approximately
$3.4\times$ and $10.9\times$ their full-method values.
Removing source-camera conditioning also increases camera errors.
These results support retaining both the target-view layout and the
source-view camera information. Figure~\ref{fig:ablation_evidence_usage} examines how evidence is used
and what happens when current observations provide insufficient support.
CGAR and EWA help preserve subject appearance and background structure.
Rendered geometry and historical retrieval address two gaps in the
available evidence: newly exposed subject surfaces and earlier scene
states beyond the rolling cache.

\begin{table*}[t]
  \centering
  \caption{Ablation results for camera-controlled video rerendering
  on DAVIS. Bold indicates the best results.}
  \label{tab:quantitative_ablation}
  \small
  \setlength{\tabcolsep}{4.0pt}
  \renewcommand{\arraystretch}{1.08}
  \begin{tabular}{lccccccc}
    \toprule
    & \multicolumn{5}{c}{VBench $\uparrow$}
    & \multicolumn{2}{c}{Camera errors $\downarrow$} \\
    \cmidrule(lr){2-6}\cmidrule(lr){7-8}
    Variant & Subj. & Bg. & Smooth. & Aesth. & Flick.
    & RotError ($^\circ$) & TransError \\
    \midrule
    w/o EWA and CGAR
    & \textbf{88.076} & 91.747 & 96.054 & 50.688 & 92.857
    & 2.0346 & 0.056937 \\
    w/o EWA
    & 88.056 & 91.706 & 96.055 & \textbf{50.720} & 92.867
    & 1.9944 & 0.056608 \\
    w/o CGAR
    & 88.055 & 91.728 & 96.072 & 50.713 & 92.891
    & 1.7828 & 0.053913 \\
    w/o Target-View Warp
    & 83.378 & 89.540 & 94.351 & 50.240 & 90.792
    & 6.1158 & 0.581847 \\
    w/o Pl\"ucker
    & 88.066 & 91.692 & 96.073 & 50.669 & 92.881
    & 1.8811 & 0.056887 \\
    \midrule
    \textbf{Full method}
    & \textbf{88.076} & \textbf{91.749} & \textbf{96.074}
    & \textbf{50.720} & \textbf{92.902}
    & \textbf{1.7823} & \textbf{0.053291} \\
    \bottomrule
  \end{tabular}
\end{table*}

%% file: sec/conclusion.tex
\section{Conclusion}
\label{sec:conclusion}

We presented \method{} (WiW), a training-free visual-evidence interface
for extending the controllability of frozen causal video world models.
WiW represents source observations, target-view projections, rendered
geometry, and generated history as camera- and time-labelled clean
visual states that the backbone reads through native self-attention.
Correspondence-guided attention routing localizes relevant source-video
evidence, while evidence-wise attention CFG regulates auxiliary
contributions using responses from the same denoising forward pass,
without additional denoising-network evaluations for guidance. On the reported DAVIS and OpenVid-1M evaluations, WiW achieves the
highest average score across seven VBench dimensions and the lowest
camera trajectory errors among the compared methods.
Qualitative results further illustrate several downstream applications through
the same interface.
These results demonstrate that constructing, selecting, and regulating
visual evidence can extend the capabilities of a pretrained world model,
enabling exploration of the dynamic world depicted in a given video.